\documentclass[conference]{IEEEtran}
\IEEEoverridecommandlockouts
\usepackage{cite}
\usepackage{amsmath,amssymb,amsfonts}
\usepackage{algorithmic}
\usepackage{graphicx}
\usepackage{textcomp}
\usepackage{xcolor}
\def\BibTeX{{\rm B\kern-.05em{\sc i\kern-.025em b}\kern-.08em
    T\kern-.1667em\lower.7ex\hbox{E}\kern-.125emX}}
\begin{document}

\title{Mimicking Associative Learning of Rats via a Neuromorphic Robot in Open Field Maze using Spatial Cell Models
}

\author{
\IEEEauthorblockN{Tianze Liu}
\IEEEauthorblockA{Department of Electrical \\and Computer Engineering \\
Michigan Technological University\\
Houghton, MI, USA \\
tianzel@mtu.edu}
\and
\IEEEauthorblockN{Md Abu Bakr Siddique}
\IEEEauthorblockA{Department of Electrical \\and Computer Engineering \\
Michigan Technological University\\
Houghton, MI, USA \\
msiddiq5@mtu.edu}
\and
\IEEEauthorblockN{Hongyu An}
\IEEEauthorblockA{Department of Electrical \\and Computer Engineering \\
Michigan Technological University\\
Houghton, MI, USA \\
hongyua@mtu.edu}
}

\maketitle

\begin{abstract}
Data-driven Artificial Intelligence (AI) approaches have exhibited remarkable prowess across various cognitive tasks using extensive training data. However, the reliance on large datasets and neural networks presents challenges such as high-power consumption and limited adaptability, particularly in SWaP-constrained applications like planetary exploration. To address these issues, we propose enhancing the autonomous capabilities of intelligent robots by emulating the associative learning observed in animals. Associative learning enables animals to adapt to their environment by memorizing concurrent events. By replicating this mechanism, neuromorphic robots can navigate dynamic environments autonomously, learning from interactions to optimize performance. This paper explores the emulation of associative learning in rodents using neuromorphic robots within open-field maze environments, leveraging insights from spatial cells such as place and grid cells. By integrating these models, we aim to enable online associative learning for spatial tasks in real-time scenarios, bridging the gap between biological spatial cognition and robotics for advancements in autonomous systems.
\end{abstract}


\section{Introduction}
Nowadays, data-driven Artificial Intelligence (AI) have demonstrated remarkable capabilities across a spectrum of cognitive tasks \cite{a1}. These capabilities are harnessed through the training process with tremendous data. Throughout the training process, the Artificial Neural Networks (ANNs) compare their outcomes with the labeled truth in the datasets. The discrepancies between the output and the truth are backpropagated into the ANNs to minimize them using the loss function, accomplished by adjusting weights through algorithms. The larger datasets and the neural networks lead to a higher accuracy \cite{a2}\cite{a3}, thereby necessitating a demand for excessive pursuit of the large scale of datasets and neural networks \cite{a2,a3} However, the continual expansion of ANNs and high dependence on labeled datasets pose several critical challenges, including high power consumption, data scarcity, and less flexibility in autonomous operating. These limitations hinder ANNs  from being feasible for Size, Weight, and Power (SWaP) restrained applications \cite{a1}\cite{a2}.  For instance, planetary rovers need to possess high adjustability and autonomous operating capabilities with minimal human intervention in environments characterized by constrained energy sources and communications \cite{a2}. 

To overcome these challenges, we enhance the autonomous operating capabilities of intelligent robots by mimicking the associative learning of animals using neuromorphic robot. Associative learning is a pervasive self-learning mechanism observed across diverse animal species. Associative learning presents the ability to adapt to the environment by interacting with their surroundings and memorizing concurrent events \cite{a8}. A classic demonstration of associative learning is an exploration of rodents in open-field maze. In the open-field maze, the rodents are presented with distinct stimuli or cues. These stimuli could be visual, sound, or a combination of sensory inputs. During the training phase, the rats learn to associate specific stimuli with favorable or unfavorable outcomes. Through repeated exposures, the rodent gradually discerns the predictive relationship between the presented cues and the associated outcomes. Thus, associative learning has the potential to empower robots with the ability to link information and experiences. In dynamic environments, such as on Mars, robots equipped with associative learning can explore unknown terrains and automatically adjust their behavior accordingly. This synergy between associative learning and adaptivity enables robots to navigate complex scenarios, learn from interactions, and autonomously optimize their performance. Several studies implemented associative learning \cite{a8,a10,a11,a13}. Nevertheless, these investigations are hindered by various limitations, including small-scale neural networks, a reliance on pure simulation rather than experimental approaches, the absence of deployment on robots for real-world scenarios testing, and so forth \cite{a13}. 

In this paper, we explore the emulation of associative learning in rats using neuromorphic robots within open-field maze environments, leveraging insights from spatial cells, including place and grid cell. Spatial cells are pivotal in associative learning. The plasticity of neural circuits, particularly in areas like the hippocampus and prefrontal cortex, supports this associative process by modifying synaptic connections based on experiences. In biological systems, this cognitive map is primarily attributed to specialized neurons known as grid and place cells within the entorhinal cortex and hippocampus. Grid cells offer a multi-scale periodic coordinate system, while place cells activate at specific locations, forming a cognitive map. This research replicates biological spatial cognition in robots, enabling complex navigation. The findings promise advancements in autonomous systems for search and rescue, planetary exploration, and GPS-denied environments, significantly enhancing robotics and AI capabilities. 

The contributions of this paper are summarized as follows:
\begin{itemize}
    \item Integrate place and grid cell models into a neuromorphic robot to conduct online associative learning for spatial tasks of rodents in real-time scenarios.
    \item Replicate associative learning in spatial memory of rodents in open-field mazes in both simulation and experimental scenarios.
\end{itemize}

\section{Computational Representation of Spatial Navigation}

In our work, we construct grid cell and place cell models that simulate spatial navigation and memory formation. Our grid cells are defined with spatial and angular parameters using vector notation, transforming positions from the physical environment to a cognitive map. The simulated grid-cell models are based on interference patterns of three two-dimensional sinusoidal gratings oriented 60° apart, consistent with previous theoretical and computational studies\cite{a14}.

Our model accurately simulates neural activity in a virtual environment based on these calculated parameters. Place cells are influenced by the regularized spatial metric provided by grid cells and exhibit firing patterns associated with specific physical locations. The activity of place cells is precisely modeled as a thresholded sum of outputs from multiple grid cells. This intricate interaction between grid and place cells ensures spatial representation and navigation with utmost precision.

In our computational framework, grid cells are defined using vector notation:
\begin{equation}
    \mathbf{G_j} = [s_j, \theta_j, \vartheta_j^1, \vartheta_j^2], \quad j \in \mathbf{Z^+},
\end{equation}
where $s_j$ is the spacing of the grid cell $G_j$, $\theta_j \in [0, \pi/3]$ is the orientation of the grid cell $G_j$, each grid cell $j$ has unique spatial and angular parameters \cite{b1}. The phases $\mathbf{\vartheta_j} = [\vartheta_j^1, \vartheta_j^2]$ are set within the interval $[0, 2\pi]$ \cite{b2}\cite{b3}, ensuring robust spatial representation.

The transformation from the place-cell frame (physical environment) to the grid-cell frame (cognitive map) is modeled by:
\begin{equation}
\begin{aligned}
    \begin{bmatrix}
    x^g_i \\
    y^g_i
    \end{bmatrix}
    &= \left(\begin{bmatrix}
    s_j \cos(\theta_j) & s_j \cos(\theta_j + \frac{\pi}{3}) \\
    s_j \sin(\theta_j) & s_j \sin(\theta_j + \frac{\pi}{3})
    \end{bmatrix}\right)^T \\
    &\quad \begin{bmatrix}
    x^p_i \\
    y^p_i
    \end{bmatrix}
    - \begin{bmatrix}
    \vartheta_j^1 \\
    \vartheta_j^2
    \end{bmatrix},
\end{aligned}
\end{equation}
where the position 
$\begin{bmatrix}
    x^p_i \\
    y^p_i
\end{bmatrix}$in the place-cell frame is mapped to the grid-cell frame 
$\begin{bmatrix}
    x^g_i \\
    y^g_i
\end{bmatrix}$, adjusting for preferred orientations and phase shifts.The neural activity for each position, indicated by the firing rate, is calculated as follows \cite{b5}:
\begin{equation}
    \varsigma_i = \arctan\left(\kappa \left(\frac{d_i}{s_j} - \zeta\right)\right).
\end{equation}
Here, $d_i$ represents the distance from the subject's position to the grid cell's preferred location. The parameter $\kappa$ is an intensity control factor, and $\zeta$ adjusts the baseline firing rate.

When the animal moves to a new environment, external cues stimulate new place cells, forming a new local place-cell frame, denoted as $C_2$. The grid-cell frame adjusts accordingly, adapting the firing field of the grid cell based on $C_2$. The initial place-cell frame $C_1$ is considered the global frame.

The position transformation is modeled by:
\begin{equation}
    \mathbf{P^{p_1}_i} = \mathbf{R_{p_1p_2}} \cdot \mathbf{P^{p_2}_i} + \mathbf{\varpi},
\end{equation}
where $\mathbf{P^{p_1}_i}$ represents the position in the initial place-cell frame, $\mathbf{P^{p_2}_i}$ is the corresponding position in the current place-cell frame, and $\mathbf{\varpi}$ is the translation vector between these two frames. The rotation matrix $\mathbf{R_{p_1p_2}}$ facilitates the transformation, adjusting the orientation between the two frames based on the rotation angle $\phi$.

The rotation matrix $\mathbf{R_{p_1p_2}}$ is described by the rotation angle $\phi$ as follows:
\begin{equation}
    \mathbf{R_{p_1p_2}} = \begin{bmatrix} \cos(\phi) & -\sin(\phi) \\ \sin(\phi) & \cos(\phi) \end{bmatrix}^T,
\end{equation}
where the superscript \(T\) denotes the transpose of the matrix. This transposition is necessary to convert the coordinate system from the current place-cell frame to the initial frame, aligning the orientation and allowing accurate spatial analysis.

To examine the grid cell's firing activity, we simulated a virtual animal path by randomly walking in different virtual environments, including a circular environment with a radius of 1.3m. The origin points of the world frame and the place-cell frame were assumed to be identical at the center of the round arena. This setup enabled us to observe and measure the grid cell's response under controlled yet dynamic conditions, mimicking natural movement within a confined space.

The grid cell used to generate the firing field is represented as:
\begin{equation}
    \mathbf{G} = [1.0, \pi/4, 0.5, 0],
\end{equation}
with hyperparameters $\zeta$ and $\kappa$ modulating the firing activity. Higher values of $\kappa$ intensify the activity around firing centers, while higher $\zeta$ values expand the firing fields, adapting the model to different environmental scales.

Our simulation explores the influence of four primary parameters—scale, orientation, kappa, and zeta—on the emergence and structure of grid cell firing fields. Adjustments in these parameters result in more pronounced activity around firing centers and expanded firing fields, depicting hexagonal patterns characteristic of grid cells. Analyzing firing rates along the directions of the grid-cell frame's basis vectors further confirms the model's precision.

\begin{figure*}[htbp]
\centering
\includegraphics[width=\textwidth]{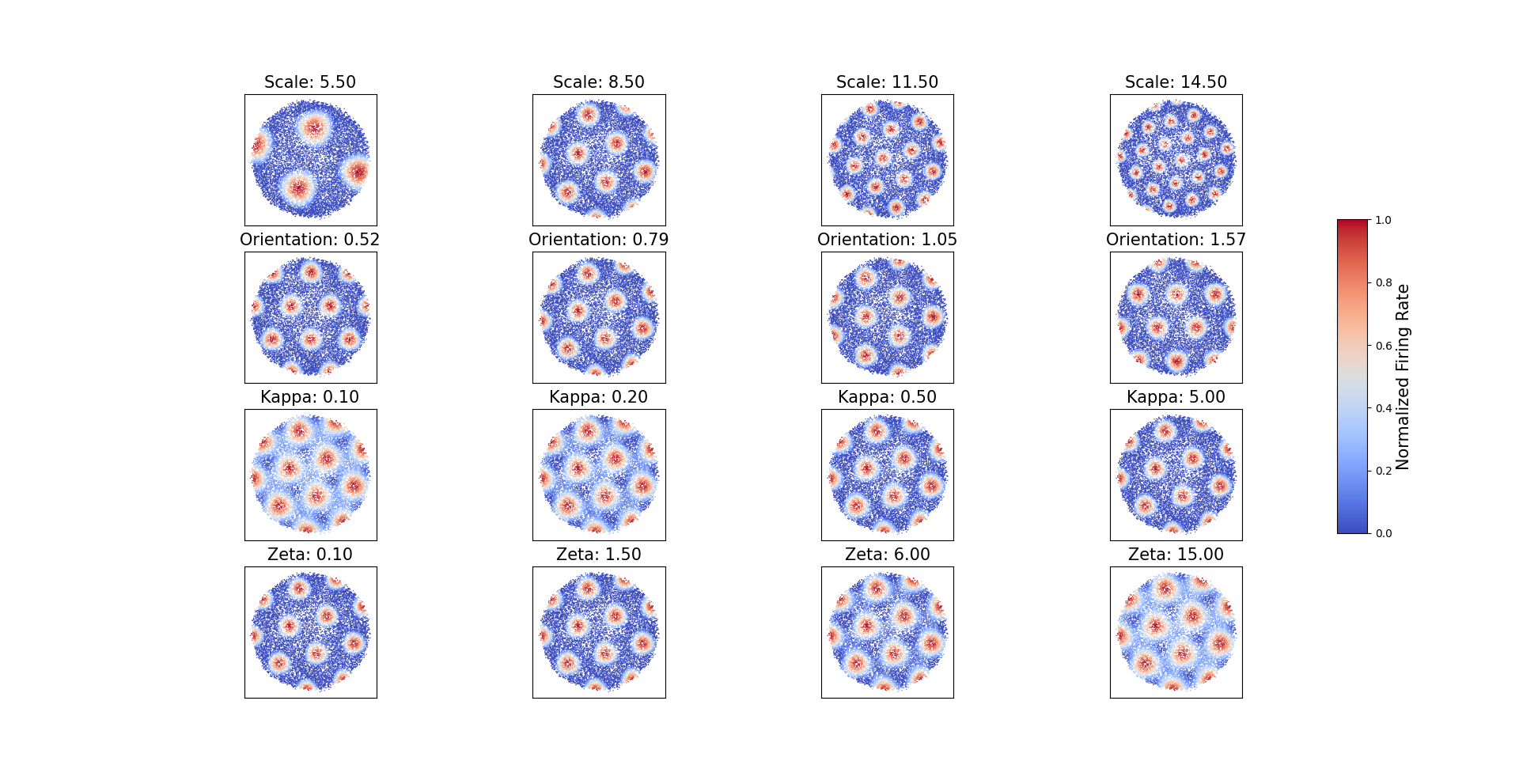} 
\caption{Simulation of grid cell firing patterns under varying parameters. Each subplot represents the grid cell activity under different scale, orientation, kappa, and zeta values, illustrating the range of firing patterns that our model can generate.}
\label{fig:grid_cell_parameters}
\end{figure*}

Figure \ref{fig:grid_cell_parameters} demonstrates the flexibility of our model by showing how varying $\kappa$ and $\zeta$ affect firing activities. Each subplot represents grid cell activity under different parameter values, illustrating the range of firing patterns our model can generate.

Experiments with multiple cells and varying parameters such as spacing, orientation, and phases led to significant changes in firing patterns, validating the model's accuracy in replicating grid cell characteristics. Increased spacing resulted in larger bumps, indicating broader spatial coverage, while adjustments in orientation and phase translated and rotated the firing fields. These results substantiate the efficacy of our model in simulating spatial representation and highlight its potential as a powerful tool for advancing the study of spatial cognition and navigation in neuroscience.

Place cells in the hippocampus are critical for spatial navigation and memory formation. These neurons exhibit firing patterns distinctly associated with specific physical locations within an environment. A place cell fires most strongly when the subject is at a particular location, known as the cell's "place field." The firing intensity of these cells decreases as the subject moves away from this central location. This unique firing characteristic ensures that each place cell responds optimally at different places, creating a spatial map within the brain.

The activity of place cells is influenced by inputs from grid cells, which provide a regularized spatial metric. The interaction between place cells and grid cells can be modeled as follows:
\begin{equation}
    \mathbf{Pc}(t) = \Theta\left(\sum_{i=1}^{N} \mathbf{Gc}_i(t)\right), 
\end{equation}
where $Pc(t)$ represents the activity function of place cells at time $t$, $\Theta$ is a step function, and $Gc_i(t)$ denotes the activity of the $i^{th}$ grid cell. This equation implies that the place cell activity is a thresholded sum of the outputs from multiple grid cells, each contributing to the overall spatial representation in the hippocampus.

To navigate and map its environment effectively, the hippocampal system utilizes visual landmarks as positional references, which are integrated into the neural representation of space through the following response function:
\begin{align}
    LP_i &= \sum \exp\left[-\frac{(d_i(t) - d_i^k(t))^2}{\partial_{d}^2} \right. \notag \\
    &\quad \left. - \frac{(\theta_i(t) - \theta_i^k(t))^2}{\partial_{\theta}^2}\right].
\end{align}

In this model, $d_i(t)$ and $\theta_i(t)$ represent the distance and angle of the $i^{th}$ landmark relative to the subject, respectively. At the same time, $\partial_d^2$ and $\partial_\theta^2$ are variance terms that adjust the sensitivity of the response to positional discrepancies. This function ensures that the spatial memory is updated accurately by adjusting for perceptual errors and discrepancies between remembered and observed landmark positions.

The integration of vibrational cues into the firing mechanisms of place cells, supported by the structured input from grid cells, offers a robust framework for understanding spatial cognition. By adapting the neural responses based on environmental stimuli and correcting for navigational errors using landmark recognition, this model underscores the dynamic nature of spatial memory and its critical role in adaptive behavior.

\section{Simulation and Experimentation of Grid Cells and Place Cells}

For simulating the grid and place cell behaviors in the neuromorphic robot, we utilized the Gazebo simulation platform, renowned for its fidelity and dynamic interaction capabilities. Gazebo provides a detailed environment that accurately models various navigation tasks and sensory scenarios, making it an indispensable tool for assessing our neuromorphic models. The simulation environment is integrated with the robot's ROS framework, ensuring high precision in simulating sensor inputs and movement responses that closely resemble real-world conditions.

Alongside Gazebo, we employ the real-time visualization tool Rviz to immediately analyze and adjust our grid and place cell models within a controlled yet adaptable virtual environment. In Gazebo, we created a circular environment to simulate the robot's Light Detection and Ranging (LiDAR)-based navigation, mirroring the real-world scenario intended for testing. Rviz complements this by providing detailed visual representations of the robot's sensory perceptions. The red points indicate the LiDAR's boundary detection, limited to 270 degrees. The green line traces the robot's path, derived from odometry data.

\begin{figure}[htbp]
\centering
\includegraphics[width=\linewidth]{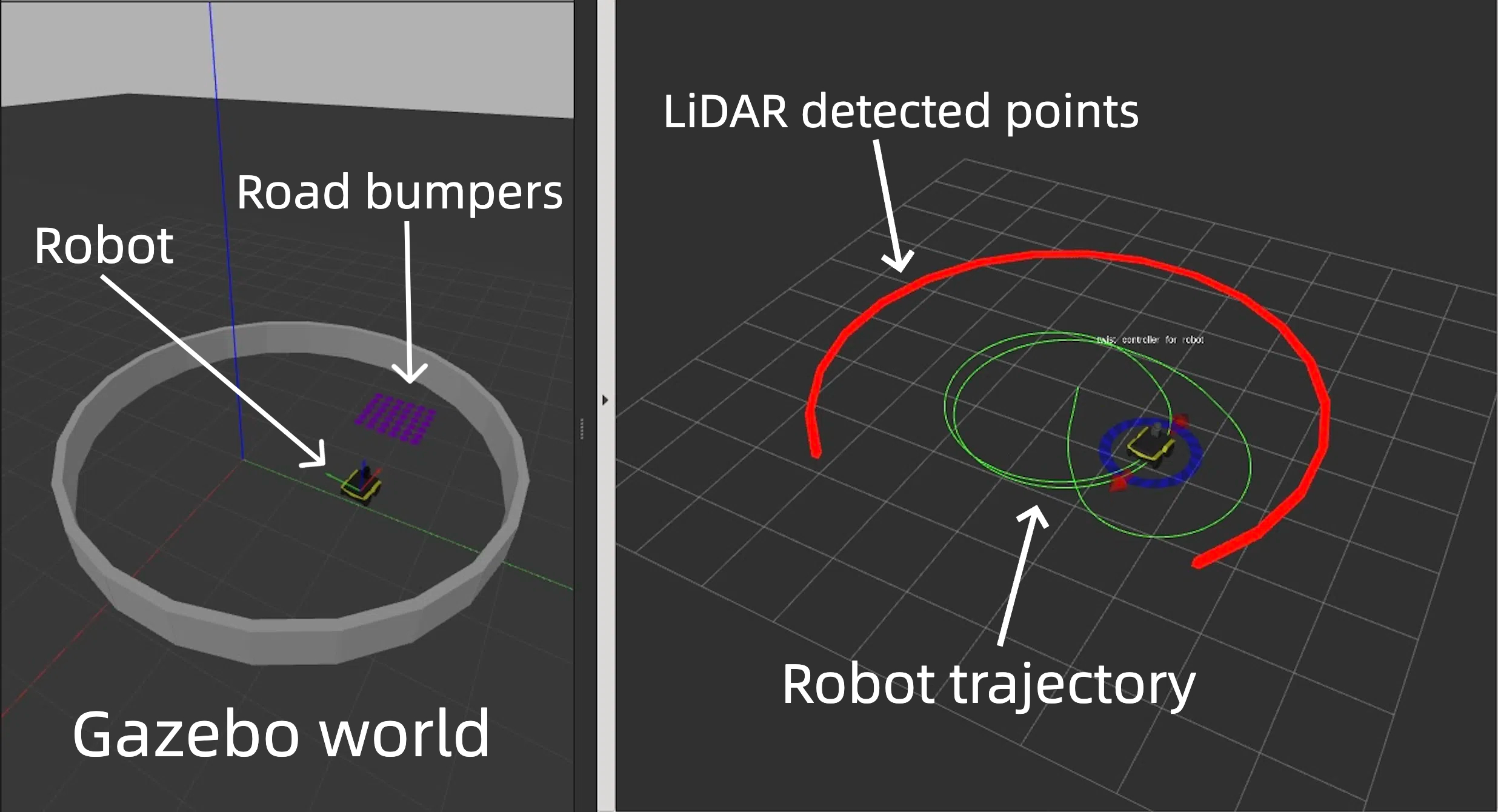}
\caption{Simulated and real-time visuals of the neuromorphic robot's path in Gazebo (left) and Rviz (right). The red line in Rviz shows LiDAR-detected boundaries with gaps indicating its 270-degree range. The green line traces the robot's path via odometry.}
\label{fig:gazebo_rviz_simulation}
\end{figure}

Figure \ref{fig:gazebo_rviz_simulation} displays our comprehensive simulation setup, showcasing the integration of Gazebo for environment modeling and Rviz for dynamic data visualization, which together form a robust platform for developing and testing advanced navigational strategies based on neuromorphic grid and place cell models.

\subsection{Integration with Neuromorphic Robot}

\subsubsection{Sensor Configuration and Data Flow}
\mbox{}\\The neuromorphic robot has an advanced sensor array, including EAI X2L LiDAR, an ORBBEC® DaBai Stereo Depth Camera, and odometers. These instruments are pivotal for the robot's obstacle avoidance, path routing, and control abilities. The Inertial Measurement Unit (IMU) detects vibrations and contributes to adjusting navigation strategies under varying physical conditions. We explain the integration process, detailing how these sensory inputs are processed within the ROS framework to influence the grid and place cell behaviors.

\subsubsection{Path Planning and Obstacle Avoidance}
\mbox{}\\The neuromorphic robot is able to achieve enhanced path planning and navigation precision through the combined data from LiDAR and IMU. The LiDAR provides detailed environmental mapping and obstacle detection, while the IMU supplies movement dynamics and orientation data. This integration refines the robot's real-time trajectory, enabling safe and efficient navigation through complex environments, optimizing routes, and maintaining stability, thereby demonstrating its practical value in the field of robotics.

\subsubsection{Vibration Analysis and Impact on Mobility}
\mbox{}\\The IMU's crucial role extends beyond contributing to the odometer system; it also enables the neuromorphic robot to analyze vibrations that affect mobility and operational efficacy. By detecting and interpreting various vibrations and their sources, the IMU helps adjust the robot's movement strategies, ensuring optimal stability and adherence to planned paths, even on uneven or dynamically changing surfaces. This analysis is precious in environments where maintaining balance and precise control over movement are challenging but necessary for successful mission outcomes.

The IMU sensors capture tri-axial acceleration data, from which the vibration is computed using the equation:
\begin{equation}
    a = \sqrt{x^2+y^2+(z-g)^2},
\end{equation}
where \( g \) approximates the gravitational acceleration constant at \( 9.81 m/s^2 \). This calculation provides a scalar magnitude of the vibrational force exerted on the robot due to irregularities in the surface texture and obstacles.

\begin{figure}[htbp]
\centering
\includegraphics[width=\linewidth]{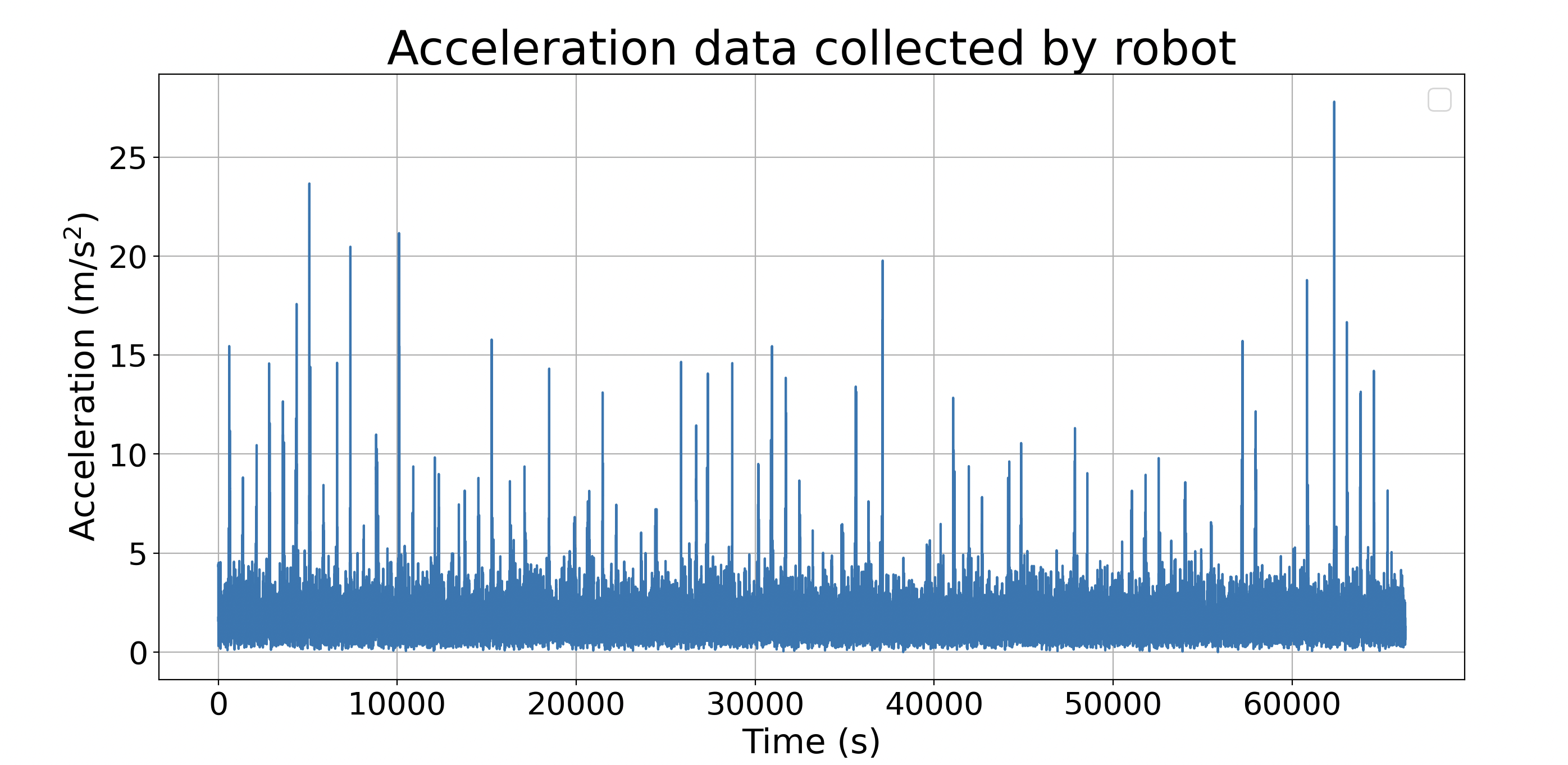}
\caption{Acceleration collected by robot in 60000 seconds, high vibration represents the robot hits the road bumpers.}
\label{fig:vibration_analysis}
\end{figure}

We visualized the data to elucidate the contrast between normal surface vibrations and those induced by bumper interactions, as shown in Figure \ref{fig:vibration_analysis}. The plot distinctly marks higher vibration, assumed to be when the robot contacts the bumpers, thus indicating a deviation from the baseline vibration levels associated with average terrain.

\subsubsection{Vibration with Color Recognition}

\begin{figure}[htbp]
\centering
\includegraphics[width=\linewidth]{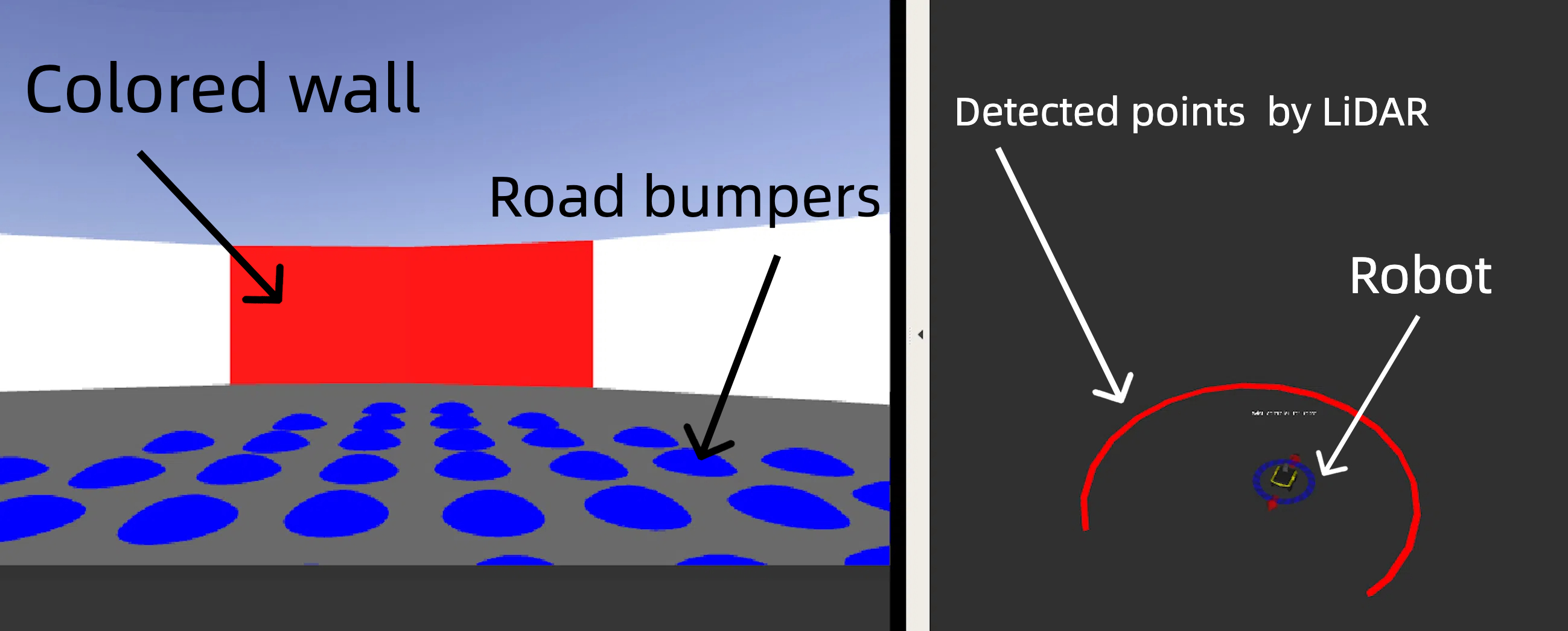}
\caption{Simulation of the robot facing a red wall, integrating visual and vibrational cues.}
\label{fig:red_wall_vibration}
\end{figure}

In Figure \ref{fig:red_wall_vibration}, the robot encounters a scenario where a red wall is presented as a visual cue and simulated ground vibrations. This test environment assesses the robot's capability to prioritize visual information in decision-making processes, remarkably when vibrations suggest an uneven area. The associative learning model adjusts the weight of visual cues in the robot's navigational method, reflecting an increased reliance on visual information when vibrations are detected.

Integrating visual and vibrational cues is crucial for developing more robust autonomous navigation systems that can operate in complex, multi-sensory environments.

\subsection{Field Testing in Real-World Scenarios}

\begin{figure}[htbp]
\centering
\includegraphics[width=\linewidth]{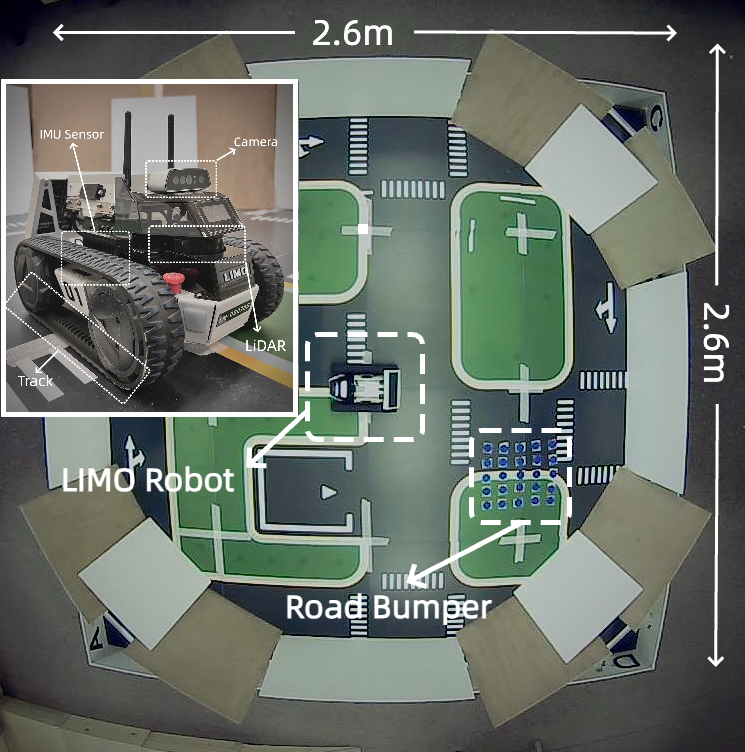}
\caption{Top view of the experimental arena used for real-world testing of the neuromorphic robot.}
\label{fig:test_environment}
\end{figure}

The real-world experiments were conducted in a controlled arena to simulate features and obstacles. As depicted in Figure \ref{fig:test_environment}, the arena spans 2.6 meters in diameter, with the neuromorphic robot starting at the center each time. Road bumpers are placed to test the robot's navigation and sensory processing capabilities in a complex setting. The layout includes various navigational challenges and is annotated with dimensions and key elements, such as the neuromorphic robot and road bumpers, to provide a scale and context.

Our real-world experimentation involved evaluating the neuromorphic robot's navigational method within an open field maze with high walls detectable by the robot's LiDAR system. This setup provided a continuous boundary simulating the operational environment that grid and place cells theorize to navigate. The testing emphasized the robot's ability to utilize its onboard sensors for orientation and navigation in environments that mimic real-world scenarios. We optimized the robot's power management systems and adapted operational strategies to address challenges such as energy constraints and data limitations encountered during these tests.

\section{Results of the Experiments}

Our experiments were designed to validate computational models by simulating the grid's navigational firing patterns and placing cells within a real robot operating in a controlled circular arena. This setup allowed us to emulate the free movement of a rat and observe the robot's behavior in a space where the biological grid cells are known to generate hexagonal firing fields.

Using the robot's navigation system, we set the primary grid cell model parameters to:
\begin{equation}
    \mathbf{G} = [8.8, \frac{\pi}{4}, 0.5, 1.2].
\end{equation}
we modulated the firing rate's range and intensity with $\zeta = 0.3$ and $\kappa = 5.0$. Concurrently, place cell models were integrated to process vibrational data, providing additional environmental context for spatial memory and navigation. The analysis of vibration , particularly when encountering road bumpers, further refined the place cell response, enhancing the robot's obstacle detection and navigation acumen.

\begin{figure*}[htbp]
\centering
\includegraphics[width=\textwidth]{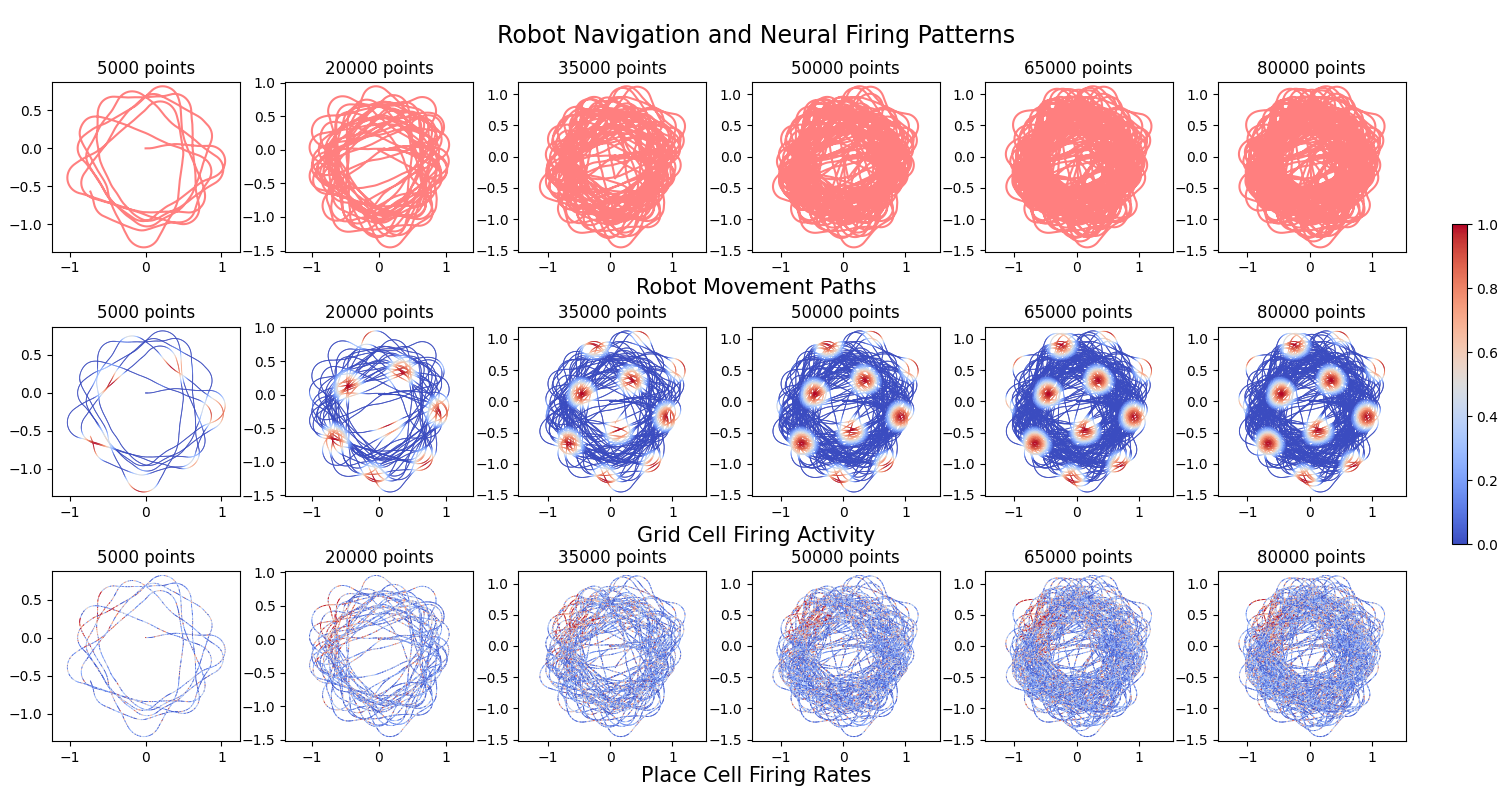}
\caption{Visualization of the robot's movement paths and the corresponding neural firing activities over increasing data points. The grid and place cell firing activities provide insight into the robot's spatial exploration and the computational model's response.}
\label{fig:rp_results}
\end{figure*}

The navigational paths and neural activity, depicted in Figure \ref{fig:rp_results}, demonstrate the robot's capability to mirror the characteristic hexagonal pattern of biological grid cells and validate the integration of place cell models informed by vibrational cues. This dual modeling approach provides robust empirical support for the accuracy of our computational navigation system. 

To enhance the robot's ability to navigate complex environments safely, we incorporated a vibration detection mechanism that triggers adaptive responses when encountering high vibration intensities. This functionality is crucial for avoiding potentially uneven areas that could impair the robot's operational integrity or hinder its path.

During experiments, the robot was programmed to alter its navigation path whenever the detected vibration exceeded a threshold of 5. This threshold was determined based on preliminary tests identifying vibration intensities typical of risky areas, such as near road bumpers or uneven terrain.

\begin{figure}[htbp]
\centering
\includegraphics[width=\linewidth]{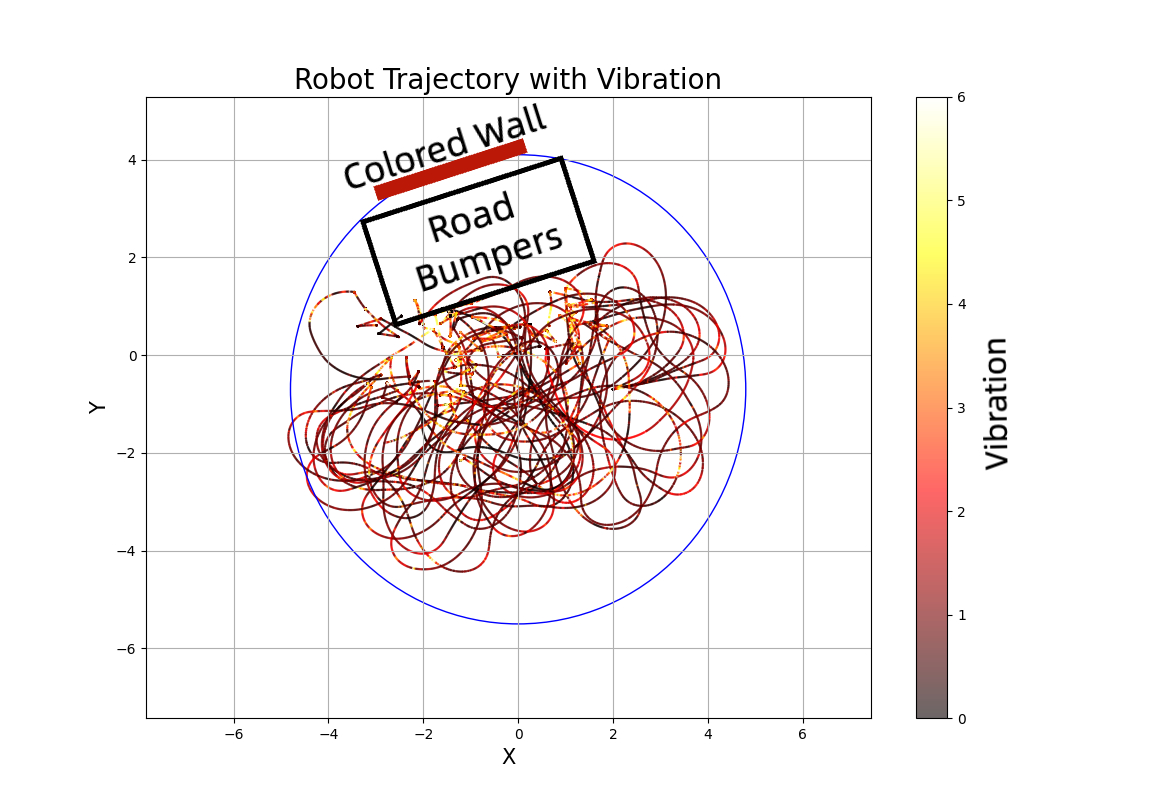}
\caption{The trajectory shows the robot avoiding areas with vibration, marked as "Road Bumpers". The avoidance behavior is triggered when vibration is detected, prompting the robot to reroute.}
\label{fig:vibration_avoidance}
\end{figure}

As depicted in Figure \ref{fig:vibration_avoidance}, the robot effectively avoids entering the high-vibration zones. The plotted trajectory illustrates how the robot approaches these zones but turns away upon reaching the vibration threshold, thus avoiding the "Road Bumpers" area. This behavior demonstrates the robot's capability to respond dynamically to sensory input and highlights the potential for such mechanisms to enhance safety and operational efficiency in autonomous navigation systems.

Our approach to enhancing the robot's navigational capabilities involved developing an associative learning algorithm that integrates visual and vibration sensory inputs. This method allowed the robot to learn from environmental interactions by associating specific colors with the vibration levels.

The algorithm enables the robot to recognize and react to environmental cues that indicate potential hazards or areas of interest. It associates visual stimuli, specifically color detection, with physical sensations such as vibrations.

During the initial phases of exploration, the robot employs its camera to detect specific colors associated with different terrain textures or obstacles. Simultaneously, the vibration sensors measure the intensity of ground vibrations, which often correlate with different surface types such as road bumpers.

As the robot encounters higher vibration intensities exceeding a predefined threshold, it is programmed to associate these intense vibrations with the visual cues at those locations. Over time, through repeated exposure and feedback, the robot's system learns to predict potential obstacles or uneven terrain based solely on visual information, even without high vibrations.

The associative learning model effectively adjusts the robot's behavior over time. Initially, the robot may try to turn around to avoid the uneven area by detecting the vibration. After it learns to associate specific colors with these vibrations, it begins to initiate avoidance upon recognizing them, anticipating and avoiding potential hazards before encountering them.

\begin{figure}[htbp]
\centering
\includegraphics[width=\linewidth]{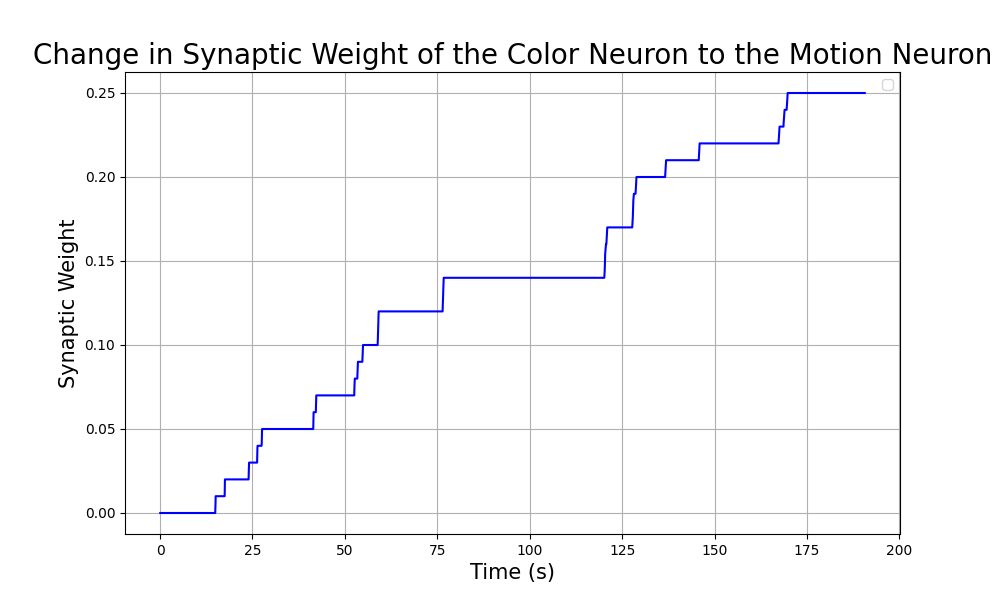}
\caption{The change of synaptic weight connecting the color neuron to the motion neuron, indicating the robot's increasing reliance on visual cues with vibration cues.}
\label{fig:color_weight_time}
\end{figure}

The effectiveness of this associative learning is quantified in Figure \ref{fig:color_weight_time}, which charts the increase in the weight assigned to color cues over time. As the robot's exposure to color-linked vibration areas increased, so did its reliance on color cues to inform its navigational decisions, demonstrating a successful integration of sensory modalities to improve autonomous navigation.

Figure 8 illustrates the change in the synaptic weight from the color neuron to the motion neuron over time. The training result demonstrates a consistent increase in the synaptic weight, which can be explained using Oja's rule. Oja's rule modifies the weight update process to stabilize Hebbian learning by introducing a normalization term. The weight change, denoted as \(\Delta w_{ij}\), is influenced by both the learning rate and the interaction between the input and the output neurons. Precisely, the weight change is calculated using the formula: 
\begin{equation} 
    \Delta w_{ij} = \eta \left( y_i x_j - y_i^2 w_{ij} \right) 
\end{equation} 

In this equation, the learning rate, represented by \(\eta\), determines the speed at which learning occurs. The term \(y_i x_j\) represents the classical Hebbian learning component, where the weight increases whenever the input and output neurons are simultaneously active. This is reflected in the graph by the upward steps, indicating periods where the activity of the color input and the motion neuron are correlated, leading to an increase in synaptic weight.

After the training phase, we conducted tests to verify the robot's ability to avoid uneven area based solely on color recognition without relying on vibration data. This test aimed to evaluate the effectiveness of the associative learning model in real-time navigation, particularly its ability to trigger avoidance behaviors based on color cues. The trajectory (black line) indicates that when the robot detects a significant amount of red pixels from the colored wall, it triggers an avoidance, demonstrating the learned behavior without needing vibration cues.

\begin{figure}[htbp]
\centering
\includegraphics[width=\linewidth]{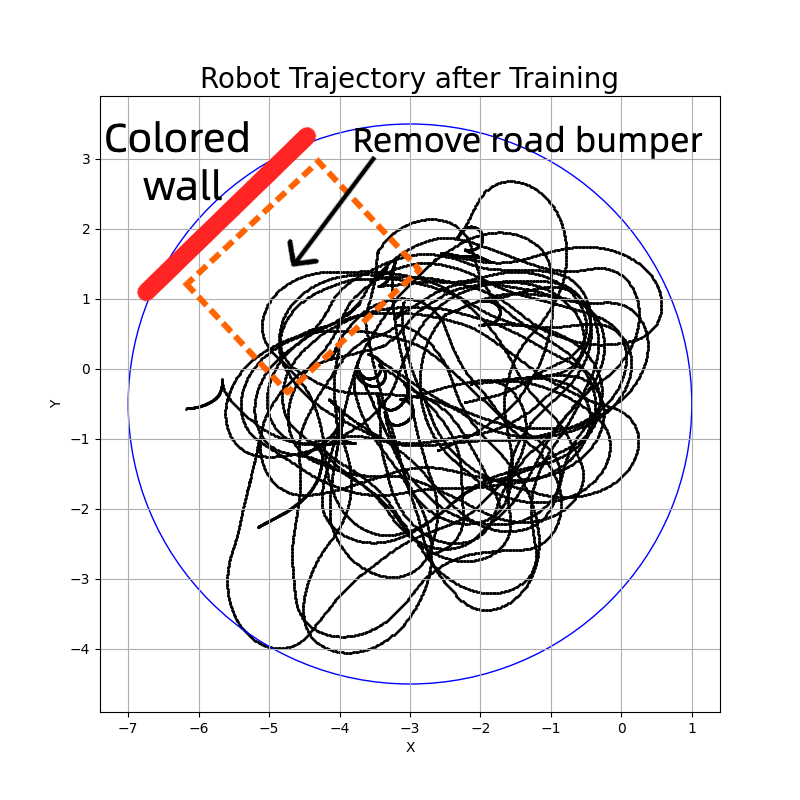}
\caption{Robot trajectory after associative learning}
\label{fig:robot_after_training}
\end{figure}

As depicted in Figure \ref{fig:robot_after_training}, the robot successfully avoids entering areas designated by the colored wall. The plotted trajectory shows that upon detecting the specified color threshold indicative of an obstacle, the robot initiates an avoidance, rerouting its path away from the obstacle. This test demonstrates that the robot can effectively use visual cues to navigate, highlighting the success of integrating associative learning into its sensory processing and decision-making framework.

This adaptive behavior is critical for autonomous navigation in complex and dynamically changing environments, where reliance on a single sensory input may not provide sufficient information for safe navigation. By integrating multiple sensory inputs and employing associative learning, the robot enhances its ability to navigate safely and efficiently, reducing the likelihood of collisions and improving its operational effectiveness in varied terrain conditions.

The implementation of this associative learning approach in the neuromorphic robot represents a significant advancement in robotics. It offers a robust method for enhancing autonomous navigation through learned environmental interactions. Integrating grid and place cell models into the neuromorphic robot's control systems has shown promising results for using neuromorphic engineering to replicate mammalian spatial mapping. This project tested the feasibility of these bio-inspired models in a robotic setting, focusing on their practical application.

Our experiments demonstrated that the robot's navigation patterns, driven by the grid and place cell models, are similar to spatial cognition in animals. The grid cell model adapted well to different environmental scales and configurations, while the place cell model updated spatial memory effectively, especially when informed by vibrational data about different terrains.

In the context of our research on neuromorphic robotics, it is pivotal to understand how our methods and results align with or diverge from other neuroscience and neuromorphic engineering studies. Table \ref{tab:neuronal_tasks} provides a comparative overview of various studies focusing on neuronal tasks, learning methods, and validation techniques. Each study is identified by its reference number, the neuronal simulation or experiment scale, and the learning methods employed.

\begin{table}[htbp]
\caption{Comparison of scale and association capability with other state-of-the-art works.}
\centering
\resizebox{\linewidth}{!}{
\renewcommand{\arraystretch}{2}
\fontsize{40pt}{60pt}\selectfont
\begin{tabular}{|c|c|c|c|c|}
\hline
\textbf{Ref} & \textbf{Neuron} & \textbf{Task} & \textbf{Learning Methods} & \textbf{Validation} \\
\hline
\cite{b6} & 6 & N/A & N/A & Simulation \\
\hline
\cite{b7} & 3 & N/A & N/A & Simulation \\
\hline
\cite{b8} & 5 & N/A & N/A & Simulation \\
\hline
\cite{b9} & 3 & N/A & N/A & Simulation \\
\hline
\cite{b10} & 3 & N/A & N/A & Simulation \\
\hline
\cite{b11} & 3 & N/A & N/A & Simulation \\
\hline
\cite{b12} & 20 & N/A & Pretraining & Simulation \\
\hline
\cite{b13}\cite{b14} & 1419 & Fear conditioning & No pretraining & Experiment \\
\hline
This work & 10 & Spatial learning and memory & Self-learning & Simulation \& Experiment \\
\hline
\end{tabular}
}
\label{tab:neuronal_tasks}
\end{table}

Despite our progress, challenges remain, such as the computational demands of simulating complex neural mechanisms and the need for enhancements to perform reliably in unpredictable conditions. However, the potential benefits of this research are significant. For example, improving computational efficiency could allow real-time processing, which is essential in dynamic environments. Further integration of learning algorithms might enhance adaptability to environmental changes. Developing multi-agent systems could lead to better collaborative mapping and task execution. Adding more types of sensory inputs might create a fuller perception system. Enhancing robustness for navigation in challenging terrains could prove invaluable in areas like disaster response or planetary exploration.

\section{Conclusion}

This study has taken some steps in neuromorphic robotics by incorporating grid and place cell models into the neuromorphic robot, improving its ability to navigate and understand its environment, akin to biological systems. Through extensive testing in simulated environments with associative learning methods, we have shown that it is possible to mimic mammalian spatial cognition with robotic systems. This work extends the capabilities of autonomous robots and enhances our knowledge of neural navigation and memory mechanisms.

\section*{Acknowledgment}

This work was supported by the Robust Intelligence program in Directorate for Computer and Information Science and Engineering (CISE) of National Science Foundation under Award Number 2245712.

\vspace{12pt}
\end{document}